%% file: arxiv.tex
\documentclass{article}

\usepackage[final, arxiv]{nips_2017}

\pdfoutput=1
\input{preamble}

\begin{document}

\input{main}

\input{acknowledgements}

\bibliography{bnns}
\bibliographystyle{iclr2017_workshop}

\include{supp_info}

\end{document}

%% file: preamble.tex
\usepackage[utf8]{inputenc} 
\usepackage[T1]{fontenc}    
\usepackage{hyperref}       
\usepackage{url}            
\usepackage{booktabs}       
\usepackage{amsfonts}       
\usepackage{nicefrac}       
\usepackage{microtype}      

\usepackage[inline, nomargin]{fixme}

\usepackage[pdftex]{graphicx}

\title{The High-Dimensional Geometry of Binary Neural Networks}

\author{
    Alexander G. Anderson\\
    Redwood Center for Theoretical Neuroscience \\
    University of California, Berkeley \\
    \texttt{aga@berkeley.edu}
\And
    Cory P. Berg\\
    Redwood Center for Theoretical Neuroscience \\
    University of California, Berkeley \\
    \texttt{cberg500@berkeley.edu}
}

%% file: main.tex
\maketitle

\begin{abstract}
Recent research has shown that
one can train a neural network with binary weights and activations at train
time by augmenting the weights with a high-precision continuous latent variable that
accumulates small changes from stochastic gradient descent.
However, there is a dearth of theoretical analysis to explain why we can
effectively capture the features in our data with binary weights and
activations.
Our main result is that the neural networks with binary weights and activations
trained using the method of Courbariaux, Hubara et al. (2016) work because of the
high-dimensional geometry of binary vectors.
In particular, the
ideal continuous vectors that extract out features in the intermediate
representations of these BNNs are well-approximated by
binary vectors in the sense that dot products are approximately preserved.
Compared to previous research that demonstrated the viability of such BNNs, our
work explains why these BNNs work in terms of the HD geometry.
Our theory serves as a foundation for understanding not only BNNs
but a variety of methods that seek to compress traditional neural networks.
Furthermore, a better understanding of multilayer binary neural networks serves as a
starting point for generalizing BNNs to other neural network architectures such
as recurrent neural networks.
\end{abstract}

\section{Introduction}


The rapidly decreasing cost of computation has driven many successes in the
field of deep learning in recent years.  Given these successes, researchers are
now considering applications of deep learning in resource limited hardware such
as neuromorphic chips, embedded devices and smart phones
(\cite{esser2016convolutional}, \cite{neftci2016neuromorphic},
\cite{andri2017yodann}). A recent success for both theoretical researchers and
industry practitioners is that traditional neural networks can be compressed
because they are highly over-parameterized.  While there has been a large
amount of experimental work dedicated to compressing neural networks (Sec.
\ref{sec:related}), we focus on the particular approach that replaces costly
32-bit floating point multiplications with cheap binary operations.  Our
analysis reveals a simple geometric picture based on the geometry of high
dimensional binary vectors that allows us to understand the successes of the
recent efforts to compress neural networks.

\begin{figure}
    \begin{center}
        \includegraphics[scale=0.8]{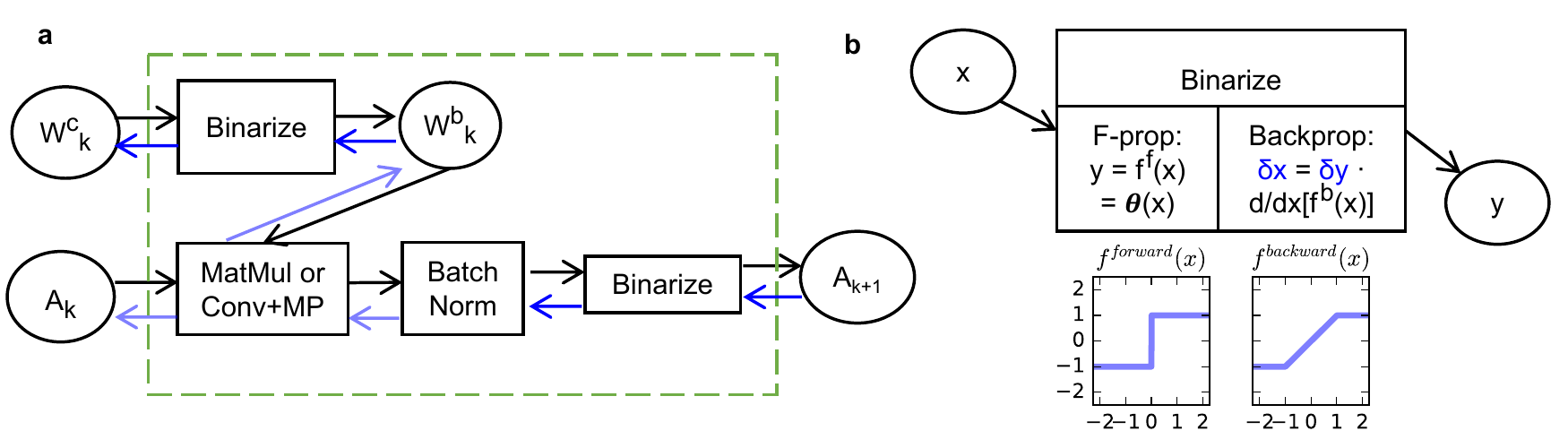}
    \end{center}
    \caption{Review of the \cite{courbariaux2016binarynet} BNN Training
        Algorithm:
        a. A binary neural network is composed of this binary convolution
        transformer (dashed green box). Each oval corresponds to a tensor and the derivative of the
        cost with respect to that tensor. Rectangles correspond to transformers that
        specify forward and backward propagation functions. Associated with each binary
        weight, $w^b$, is a continuous weight, $w^c$, that is used to
        accumulate gradients. $k$ denotes the $k$th layer of the network.
        b. Each binarize transformer has a forward function and a backward
        function. The forward function simply binarizes the inputs. In the backward propagation
        step, one normally computes the derivative of the cost with respect to
        the input of a transformer
        via the Jacobian of the forward function and the derivative of
        the cost with respect to the output of that transformer ($\delta u \equiv dC/du$ where $C$
        is the cost function used to train the network). Since the binarize function is
        non-differentiable, they use a smoothed version of the forward function
        for the backward function (in particular, the
        straight-through estimator of \cite{bengio2013estimating}).
    }
    \label{fig:alg_review}
\end{figure}

Recent work by \cite{courbariaux2016binarynet} and
\cite{hubara2016quantized} has shown
that one can efficiently train neural networks with binary weights and
activations that have similar performance to their continuous counterparts.
They demonstrate that such BNNs execute 7 times faster using a
dedicated GPU kernel at test time.  Furthermore, they argue that such BNNs
require at least a factor of 32 fewer memory accesses at test time that
should result in an even larger energy savings.  There are
two key ideas in their papers (Fig. \ref{fig:alg_review}).
First, they associate a continuous weight, $w^c$, with each binary weight, $w^b$,
that accumulates small changes from stochastic gradient descent.
Second, they replace the non-differentiable binarize function ($\theta(x) = 1$
if $x > 0$ and $-1$ otherwise) with a continuous one during backpropagation.
These modifications allow them to train neural networks that have binary weights
and activations with stochastic gradient descent.
While their work has shown how to train such networks, the existence of neural
networks with binary weights and activations needs to be reconciled with
previous work that has sought to understand weight matrices as extracting out
continuous features in data (e.g.  \cite{zeiler2014visualizing}).
Summary of contributions:

\begin{enumerate}
    \item Angle Preservation Property: We demonstrate that binarization
        approximately preserves the direction of high dimensional vectors. In
        particular, we show that  the angle between a random vector (from a standard normal
        distribution) and its binarized version converges to $\arccos
        \sqrt{2/\pi}\approx 37^\circ$ as the dimension of the vector goes to
        infinity.
        This angle is an exceedingly small angle in high dimensions.
        Furthermore, we show that this property is present in the weight
        vectors of a network trained using the method of
        \cite{courbariaux2016binarynet}.

    \item Dot Product Preservation Property:
        We show that the
        batch normalized weight-activation dot products, an important
        intermediate quantity in these BNNs, are
        approximately preserved under the binarization of the weight vectors.
        Relatedly, we argue that the continuous weights in the
        \cite{courbariaux2016binarynet} method aren't just a learning artifact
        - they correspond to continuous weights trained using an estimator of
        the true gradient. Finally, we argue that this dot product preservation
        property is a sufficient condition for the modified learning dynamics
        to approximate the true learning dynamics that would
        train the continuous weights.

    \item Generalized Binarization Transformation:
        We show that the computations done by the first layer of the network
        on CIFAR10 are fundamentally different than the computations being done
        in the rest of the network because the high variance principal components are not
        randomly oriented relative to the binarization.
        Thus we recommend an architecture that uses a continuous convolution
        for the first layer to embed the image in a high dimensional binary
        space, after which it can be manipulated with cheap binary operations.
        Furthermore, we hypothesize that a GBT (rotate, binarize, rotate
        back) will be useful for dealing with low dimensional data embedded in
        a HD space that is not randomly oriented relative to the axes of
        binarization.
\end{enumerate}

\section{Related Work}
\label{sec:related}

Neural networks that achieve good performance on tasks such as
IMAGENET object recognition are highly computationally intensive. For instance, AlexNet has
61 million parameters and executes 1.5 billion operations to classify one 224 by 224 image
(30 thousand operations/pixel)
(\cite{rastegari2016xnor}). There has been a substantial amount of work to reduce
this computational cost for embedded applications.

First, there are a variety of approaches that seek to compress
pre-trained networks. \cite{kim2015compression} uses a Tucker decomposition of
the kernel tensor and fine tunes the network afterwards.
\cite{han2015learning} train a network, prune low magnitude connections, and
retrain.
\cite{han2015deep} extend their previous work to additionally include a weight
sharing quantization step and Huffman coding of the weights.

Second, researchers have sought to train networks
using either low precision floating point numbers or fixed point numbers, which
allow for cheaper multiplications (\cite{courbariaux2014training},
\cite{gupta2015deep}, \cite{judd2015reduced}, \cite{gysel2016hardware},
\cite{lin2016fixed},
\cite{lai2017deep}).

Third, there are numerous works on training networks that have quantized weights
and or activations.
Classically, \cite{bengio2013estimating} looked
at a variety of estimators for the gradient through a stochastic binary unit.
\cite{courbariaux2015binaryconnect} trains networks with binary weights, and
then later with binary weights and activations (\cite{courbariaux2016binarynet}).
\cite{rastegari2016xnor} replace a continuous weight matrix with a scalar times
a binary matrix (and have a similar approximation for weight activation dot
products).
\cite{kim2016bitwise} train a network with weights restricted in the range $-1$
to $1$ and then use a noisy backpropagation scheme train a network with
binary weights and activations.
\cite{alemdar2016ternary}, \cite{li2016ternary} and \cite{zhu2016trained} focus
on networks with ternary weights.
Further work seeks to quantize the weights and activations in neural networks
to an arbitrary number of bits
(\cite{zhou2016dorefa}, \cite{hubara2016quantized}).
\cite{zhou2017incremental} use weights and activations that are zero or powers of
two.
\cite{lin2015neural} and \cite{zhou2016dorefa} quantize backpropagation in
addition to the forward propagation.

Beyond merely seeking to compress neural networks, there is a variety of papers
that seek to analyze the internal representations of neural networks.
\cite{agrawal2014analyzing} found that feature magnitudes in higher layers do
not matter (e.g. binarizing features barely changes classification
performance).  \cite{merolla2016deep} analyze the robustness of neural network
representations to a collection of different distortions.
\cite{dosovitskiy2016inverting} observe that binarizing features in
intermediate layers of a CNN and then using backpropagation to find an image
with those features leads to relatively little distortion of the image compared
to dropping out features. These works naturally lead into our work where we are
seeking to better understand the representations in neural networks based on
the geometry of HD binary vectors.

\section{Theory and Experiments}

In this section, we outline two theoretical predictions and then verify them
experimentally. We train a binary neural network
on CIFAR-10 (same learning algorithm and architecture as in
\cite{courbariaux2016binarynet}).  This convolutional neural network has six
layers of convolutions, all of which have a 3 by 3 spatial kernel.
The number of feature maps in each layer are 128, 128, 256, 256,
512, 512. After the second, fourth, and sixth convolutions, we do a 2 by 2
max pooling operation. Then we have two fully connected layers with $1024$ units
each.  Each layer has a batch norm layer in between.  We note that the
dimension of the weight vector in consideration (i.e. convolution converted to
a matrix multiply) is the patch size ($=3*3=9$) times the number of channels.
We also carried out experiments using MNIST and got similar
results.

\subsection{Preservation of Direction During Binarization}

\begin{figure}
    \begin{center}
        \includegraphics[scale=0.8]{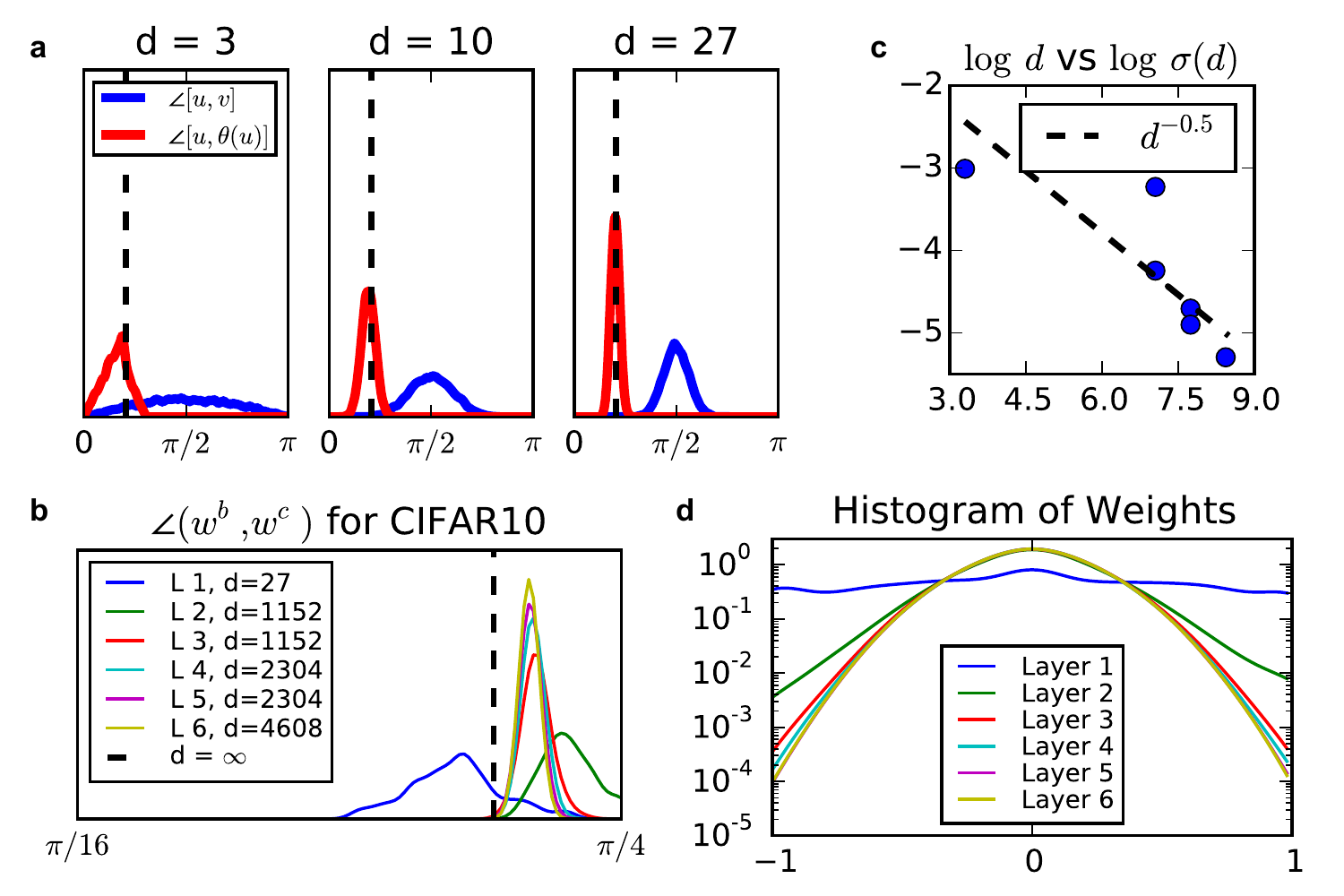}
    \end{center}
    \caption{
        Binarization of Random Vectors Approximately Preserves their Direction:
        (a) Distribution of angles between two random vectors, and between a vector and
        its binarized version, for a rotationally invariant distribution.
        Looking at the angle between a random vector of dimension $d$ and its binarized version,
        we get the red curves. We note that the distribution is peaked near the
        $d\rightarrow\infty$ limit of $\arccos
        \sqrt{2/\pi}\approx 37^\circ$ (SI, Sec. 1). Is
        $37^\circ$ a small angle or a large angle? In order to think about
        that, consider the distribution of angles between two random vectors
        (blue curves).
        We see that for low dimensions, the angle between a vector and
        its binarized version has a high probability of being similar to a pair
        of two random vectors.  However, when we get to a moderately high
        dimension, we see that the red and blue curves are well separated.
        (b) Angle distribution between continuous and binary weight vectors by
        layer for a binary CNN trained on CIFAR-10. For the higher layers, we
        see a relatively close correspondence with the theory, but
        with a systematic deviation towards slightly larger angles. $d$ is the
        dimension of the filters at each layer.
        (c) Standard deviations of the angle
        distributions from (b) by layer. We see a correspondence to the
        theoretical expectation that
        standard deviations of each of the angle distributions scales as
        $d^{-0.5}$ (SI, Sec. 1).
        (d) Histogram of the components of the continuous weights at each
        layer. We note that this distribution is
        approximately Gaussian for all but the first layer.
        Furthermore, we note that there is a high density of weights near zero, which is the
        threshold for the binarization function.
    }
    \label{fig:random_binarization}
\end{figure}

In the hyperdimensional computing theory of \cite{kanerva2009hyperdimensional},
one of the key ideas is that two random, high-dimensional vectors of dimension
$d$ whose entries are chosen uniformly from
the set $\{-1, 1\}$ are approximately orthogonal
(by the central limit theorem, the cosine angle between two such random vectors is
normally distributed with $\mu=0$ and $\sigma\sim 1/\sqrt{d}$ ($\cos\theta
\approx 0 \rightarrow \theta \approx \frac{\pi}{2}$)).
We apply this approach of
analyzing the geometry (i.e. angle distributions) of high-dimensional vectors
to binary vectors.  As a null distribution, we use the standard normal
distribution, which is rotationally invariant, to generate our vectors.
In moderately high dimensions, binarizing a random vector changes
its direction
by a small amount relative to the angle between two random vectors.  This is
contrary to our low-dimensional intuition that is guided by the fact that the
angle between two random 2D vectors is uniformly distributed (Fig.
\ref{fig:random_binarization}).

In order to test the applicability of our theory of Gaussian random vectors to
real neural networks, we
train a multilayer binary CNN on CIFAR10 (using the
\cite{courbariaux2016binarynet} method) and look at the weight vectors
\footnote{If we write each convolution as the matrix
multiplication $Wx$ where $x$ is a column vector, then the weight vectors are the rows of
$W$.} in that
trained network.  We see a close correspondence between the experimental results
and our theory for the angles between the binary
and continuous weights (Fig.  \ref{fig:random_binarization}).
We note that there is a small but systematic deviation from the theory towards
larger angles for the higher layers of the network.

\subsection{Dot Product Preservation as a Sufficient Condition for Sensible
Learning Dynamics}
One reasonable question to ask is: are these so-called
'continuous weights' just a learning artifact without a clear correspondence to
the binary weights?
While we know that $w^b = \theta(w^c)$, there are many continuous weights that
map onto a particular binary weight vector. Which one do we find when we apply
the method of \cite{courbariaux2016binarynet}? As we discuss below, we get the
continuous weight that preserves the dot products with the activations.
The key to our analysis is to focus on the transformers in our network whose
forward and backward propagation functions are not related in the way
that they would normally be related in typical gradient descent.

We show that the modified gradient that we are using
can be viewed as an estimator of the true gradient that would be used to train
the continuous weights in traditional backpropagation.
Furthermore, we show that a sufficient property for this estimator to be a good
one is that the dot products of the activations with the pre-binarized and
post-binarized weights are proportional.

Suppose we have a neural network where we allocate two tensors, $u$, and $v$
(and the associated derivatives of the cost with respect to those tensors,
$\delta u$ and $\delta v$).
Suppose that the loss as a function of $v$ is $L(x)|_{x=v}$.  Further,
there are two potential forward propagation functions, $f$, and $g$. If we
trained our network under normal conditions using $g$ as the forward
propagation function, then we would compute

$$v\leftarrow g(u)   \qquad \delta v \leftarrow L'(x=v=g(u)) \qquad \delta u
\leftarrow \delta v \cdot g'(u)$$

In the modified backpropagation scheme, we compute

$$v \leftarrow f(u) \qquad \delta v = L'(x=v=f(u)) \qquad \delta u \leftarrow
\delta v \cdot g'(u)$$
A sufficient condition for the updates $\delta u$ to be the same is $L'(x=f(u))
\sim L'(x=g(u))$ where $a \sim b$ means that the vector $a$ is a scalar times the
vector $b$.

Now we specialize this argument to the binarize block that binarizes the
weights in our networks. Here, $u$ is the continuous weight, $w^c$, $f(u)$ is the pointwise
binarize function, $g(u)$ is the identity function \footnote{For the
weights, $g$ as in Fig. \ref{fig:alg_review} is the identity function because
the $w^c$'s are clipped to be in $[-1,1]$}, and $L$ is the loss of the
network as a function of the weights in a particular layer. Given our
architecture, we can write $L(x) = M(a \cdot x)$ where $a$ are the activations
corresponding to that layer ($a$ is binary for all except the first layer) and
$M$ is the loss as a function of the weight-activation dot products. Then
$L'(x) = M'(a \cdot x) \odot a$ where $\odot$ denotes a pointwise multiply.
Thus the sufficient condition is
$M'(a\cdot w^b) \sim M'(a\cdot w^c)$.  Since the dot products are followed
by a batch normalization, $M(k\vec{x}) = M(\vec{x})\rightarrow M'(\vec{x}) = k
M'(k\vec{x})$.  Therefore, it is sufficient that

$$a\cdot w^b \sim a\cdot w^c$$

We call this final relation the Dot Product Preservation (DPP) property.
In summary, the learning dynamics where we use $g$ for the forward and backward passes
(i.e. training the network with continuous weights) is approximately equivalent
to the modified learning dynamics ($f$ on the forward pass, and $g$ on the
backward pass) when we have the DPP property.

We also come at this problem from another direction. In SI, Sec.
2 we work out the learning dynamics of the modified
backprop scheme in the case of a
one layer neural network that seeks to do regression (this ends up being linear
regression with binary weights). In this case, the learning dynamics for the
weights end up being $\Delta w^c \sim C_{yx} - \theta(w^c) C_{xx}$ where
$C_{yx}$ is the input-output correlation matrix and $C_{xx}$ is the input
covariance matrix. Since $\theta$ forces the weight matrix to be binary, this
equation cannot be satisfied exactly in general conditions. Specializing to the
case of an identity input covariance matrix, we
show that $E(\theta(w^c)) = C_{yx}$. Intuitively, the entries
of the weight matrix oscillate between $+1$ and $-1$ in the correct proportion
in order to get the weight matrix correct in expectation. In high dimensions,
these are likely to be out of phase, leading to a low variance estimator.

\begin{figure}
    \begin{center}
        \includegraphics[scale=0.75]{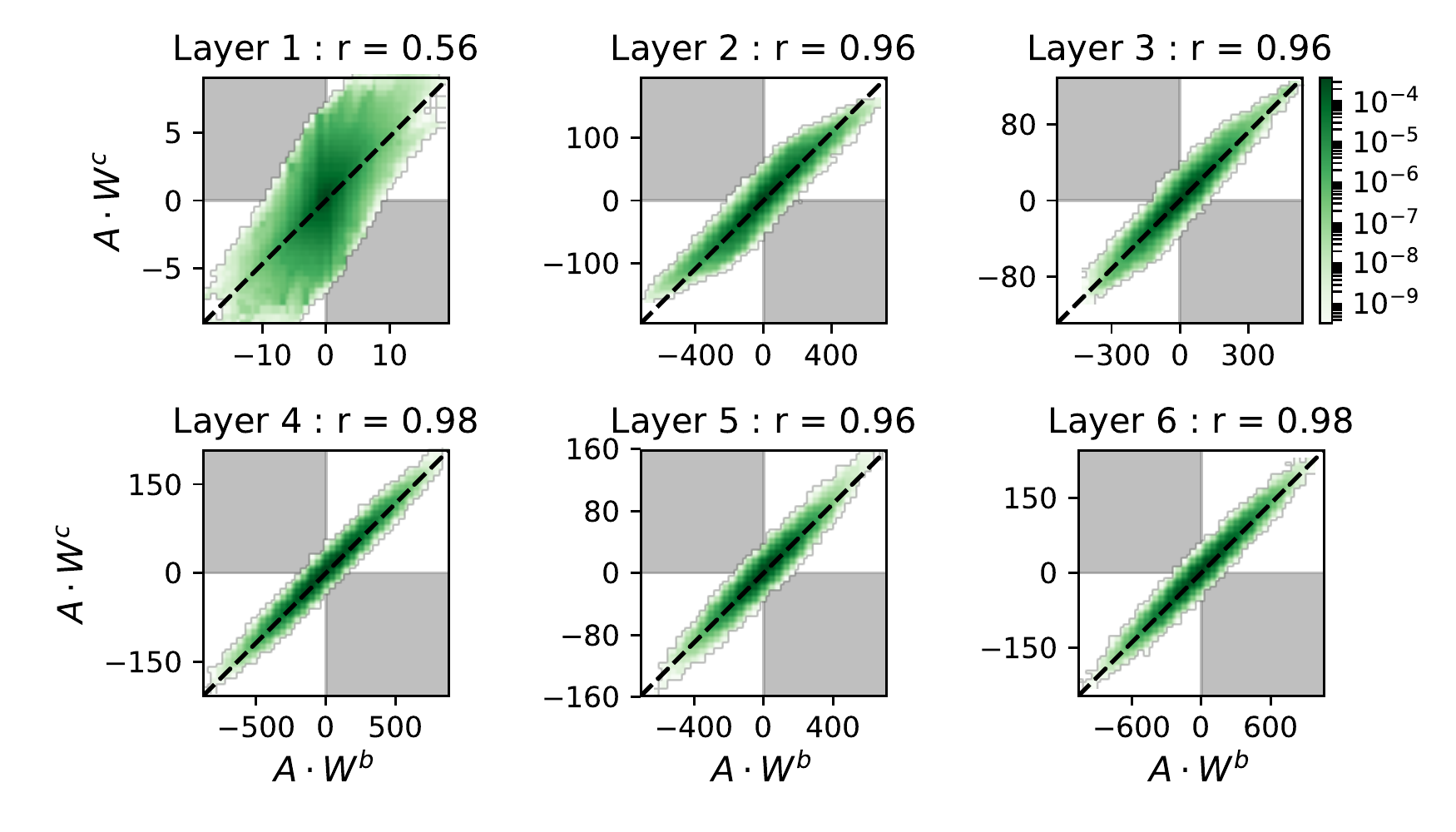}
    \end{center}
    \caption{
        Binarization Preserves Dot Products: In this figure, we verify our
        hypothesis that binarization approximately preserves the dot-products
        that the network uses for computations. We train a convolutional neural
        network on CIFAR-10.  Each figure shows a 2d histogram of the dot
        products between the binarized
        weights and the activations (horizontal axis) and the dot products
        between the continuous
        weights and the activations (vertical axis). Note the logarithmic
        color scaling. We see that these dot products are highly correlated,
        verifying our theory ($r$ is the Pearson correlation coefficient). Note
        they may differ up to a scaling constant due to the subsequent batch
        norm layer. The top left figure (labeled as Layer 1) corresponds to the
        input and the first convolution. Note that the correlation is weaker
        in the first layer. The shaded quadrants correspond to dot products
        where the sign changes when replacing the binary weights with the
        continuous weights. Notice that for all but the first layer, a very
        small fraction of the dot products lie in these off diagonal quadrants.
    }
    \label{fig:preserve}
\end{figure}

Indeed, in our numerical experiments on CIFAR10, we see that the dot products
of the activations with the pre-binarization and post-binarization weights are
highly correlated (Fig.  \ref{fig:preserve}).
Likewise, we verify a second relation that corresponds to ablating the other
instance of binarize transformer in the network, the transformer that binarizes
the activations : $w^b \cdot a^c \sim w^b \cdot a^b$ where $a^c$ denotes the
pre-binarized (post-batch norm) activations (Fig. \ref{fig:controls}).  For the
practitioner, we recommend checking the
DPP property in order to assess the areas in which the network's performance
is being degraded by the compression of the weights or activations.

Impact on Classification: As we've argued, the quantity that the network cares
about, the batch normalized weight-activation dot products, is preserved under
binarization of the weights. It is also natural to ask to what extent the
classification performance depends on the binarization of the weights. In our
experiments on CIFAR10, if we remove the binarization of the weights on all of
the convolutional layers, the classification performance drops by only $3$
percent relative to the original network. Looking at each layer individually,
we see that removing the weight binarization for the first layer accounts for
this entire percentage, and removing the binarization of the weights for each
other layer causes no degradation in performance.
We note that removing the binarization of the activations unsurprisingly has a
substantial impact on the classification performance because that
removes the main non-linearity of the network.

\subsection{Permutation of Activations Reveals Fundamental Difference Between
First Layer and Subsequent Layers}

\begin{figure}
    \begin{center}
        \includegraphics[scale=0.35]{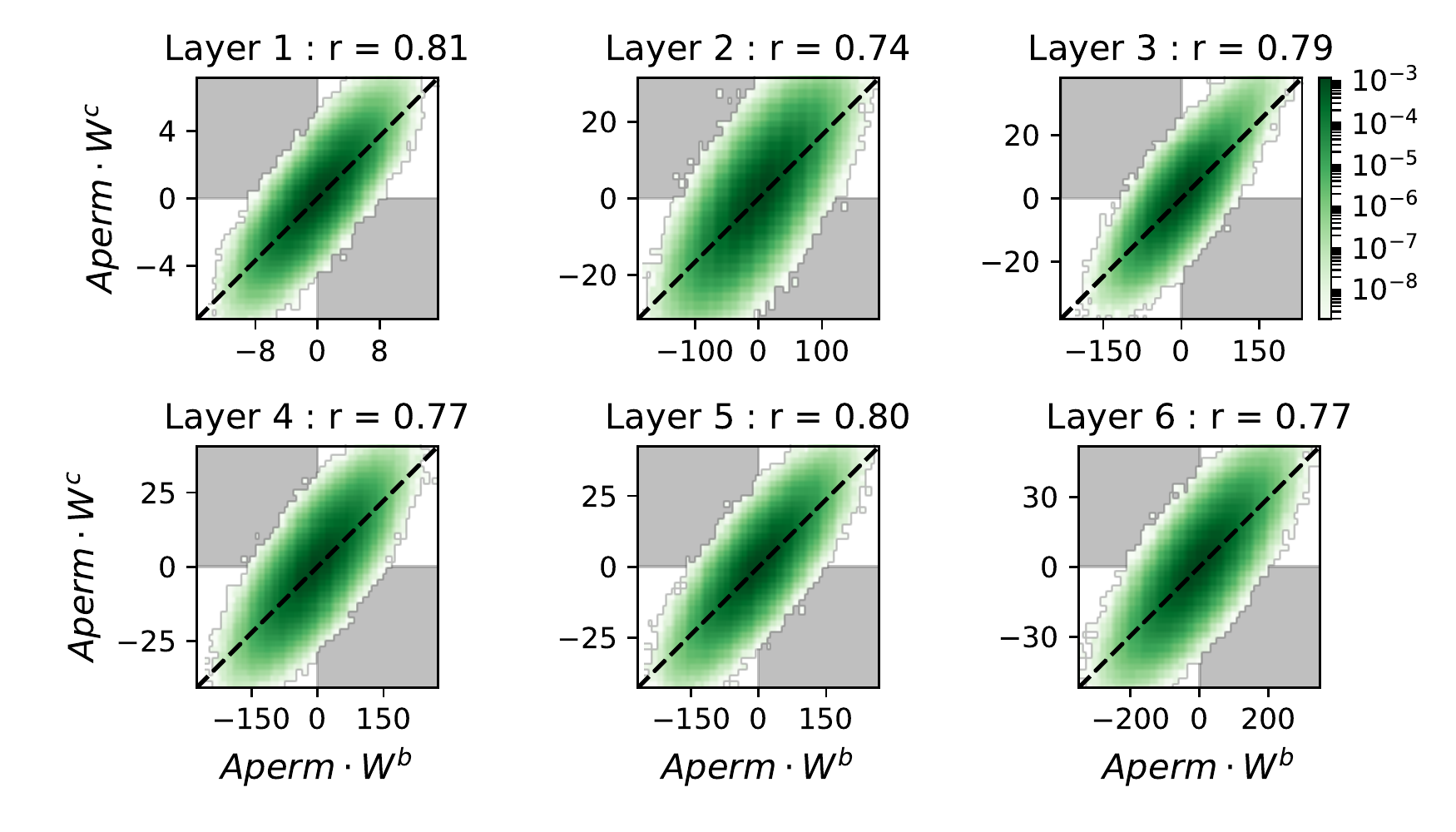}
        \includegraphics[scale=0.35]{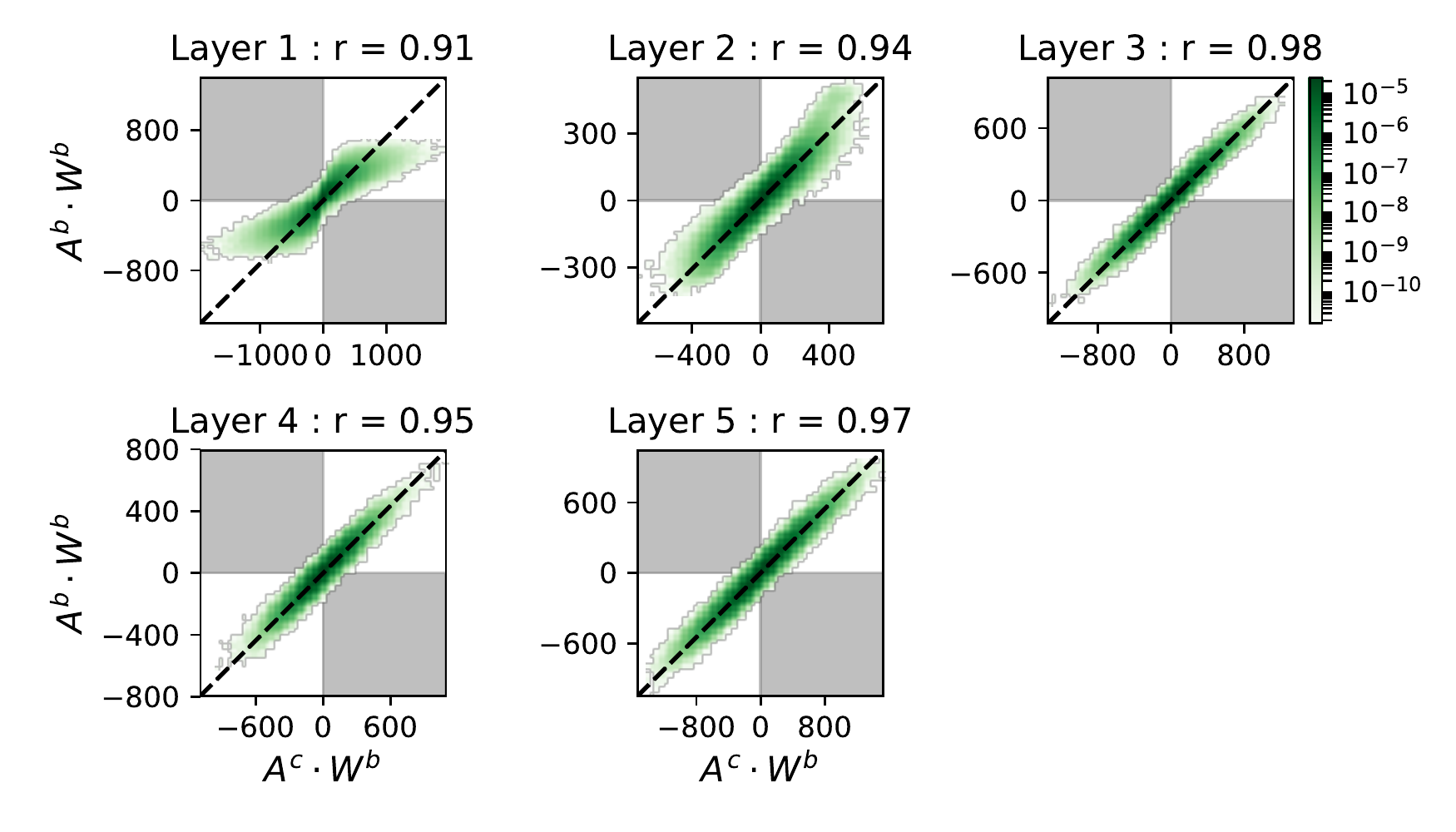}
    \end{center}
    \caption{
        Left: Permuting the activations shows that the correlations observed in Fig.
        \ref{fig:preserve} are not merely due to correlations between the
        binary and continuous weight vectors. The correlations are due to these
        weight vectors corresponding to important directions in the data.
        Right: Activation Binarization Preserves Dot Products: Each figure shows
        a 2d histogram of the dot products between the binarized weights and binarized
        activations (vertical axis) and post-batch norm (but pre activation
        binarization) activations (horizontal axis). Again, we see that the
        binarization transformer does little to corrupt the dot products
        between weights and activations.
}
    \label{fig:controls}
\end{figure}

Looking at the correlations in Fig. \ref{fig:preserve}, we see that the first
layer has a much smaller dot product correlation than the other layers.
In order to understand this observation better, we investigate the different
factors that lead to the dot product correlation.
For instance, it could be the case that the correlation between the two dot
products is due to the two weight vectors being closely aligned. Another
explanation is that the weight vectors are well-aligned with the informative
directions in the data.
To study this, we apply a random permutation to the
activations in order to generate a distribution with the same marginal
statistics as the original data but independent joint statistics.
Such a
transformation gives us a distribution with a correlation equal to the
normalized dot product of the weight vectors (SI Sec. 3).
As we can see in Fig.  \ref{fig:controls}, the correlations for the higher
layers decrease substantially but the correlation in the first layer
\textit{increases} (for the first layer, the shuffling operation randomly permutes the
pixels in the image).
Thus we demonstrate that the binary weight vectors in the first layer are not
well-aligned with the continuous weight vectors relative to the input data.

We hypothesize that the core issue at play is that the input data
is not randomly oriented relative to the axes of binarization.
In order to be clear on what we mean by the axes of binarization, first
consider the Generalized Binarization
Transformation (GBT): $$\theta_R(x) = R^{T} \theta(R x)$$ where
$x$ is a column vector, $R$ is a rotation matrix, and $\theta$ is the pointwise
binarization function from before. We call the rows of $R$ the axes of binarization. If
$R$ is the identity matrix, then we reduce back to our original binarization
function and the axes of binarization are just the canonical basis vectors
$(..., 0, 1, 0, ... )$.
Consider the 27 dimensional input to the first set of convolutions in our
network: 3 color channels of a 3
by 3 patch of an image from CIFAR10 with the mean removed.
3 PCs capture 90 percent of the variance of this data and 4 PCs
capture 94.5 percent of the variance.
Furthermore, these PCs aren't randomly oriented relative to the binarization axes.
For instance, the first two PCs are spatially uniform colors.
More generally, natural images (such as those in IMAGENET) will have the same
issue.
Translation invariance of the pixel covariance matrix implies that the
principal components are the filters of the 2D fourier transform. Scale invariance
implies a $1/f^2$ power spectrum, which results in the largest PCs
corresponding to low frequencies.


Stepping back, this control gives us important insight: the first layer is
fundamentally different from the other layers due to the non-random orientation
of the data relative to the axes of binarization.
Practically speaking, we have two recommendations.
First, we recommend an architecture that
uses a set of continuous convolutional weights to embed images in a
high-dimensional binary space, after which it can be manipulated efficiently
using binary vectors.
While there isn't a large accuracy degradation on CIFAR10, these observations
are going to be more important on datasets with larger images such as IMAGENET.
We note that this theoretically grounded recommendation is
consistent with previous empirical work.  \cite{han2015learning} find that
compressing the first set of convolutional weights of a particular layer by the
same fraction has the highest impact on performance if done on the first layer.
\cite{zhou2016dorefa} find that accuracy degrades by about 0.5 to 1
percent on SHVN when quantizing the first layer weights.
Second, we recommend experimenting with a GBT
where the rotation is chosen so that it can be computed
efficiently.
This solves the problem of low-dimensional data embedded
in a high dimensional space that is not randomly oriented relative to the
binarization function.



\section{Conclusion}

Neural networks with binary weights and activations have similar performance to
their continuous counterparts with substantially reduced execution time and
power usage.
We provide an experimentally verified theory for understanding how one can
get away with such a massive reduction in precision based on the geometry of
HD vectors.
First, we show that binarization of high-dimensional vectors preserves their
direction in the sense that the angle between a random vector and its binarized
version is much smaller than the angle between two random vectors (Angle
Preservation Property).
Second, we take the perspective of the network and show that
binarization approximately preserves weight-activation dot products (Dot
Product Preservation Property).
More generally, when using a network compression technique, we recommend
looking at the weight activation dot product histograms as a heuristic to help
localize the layers that are most responsible for performance degradation.
Third, we discuss the impacts of the low effective dimensionality on the first
layer of the network and recommend either using continuous weights for the first
layer or a Generalized Binarization Transformation. Such a transformation may
be useful for architectures like LSTMs where the update for the hidden state
declares a particular set of axes to be important (e.g. by taking the pointwise
multiply of the forget gates with the cell state).
More broadly speaking, our theory is useful for analyzing a
variety of neural network compression techinques that
transform the weights, activations or both to reduce the execution cost
without degrading performance.



%% file: acknowledgements.tex
\subsubsection*{Acknowledgments}
The authors would like to thank Bruno Olshausen, Urs K\"oster, Spencer Kent,
Eric Dodds, Dylan Paiton, and members of the
Redwood Center for useful feedback on this work.
This material is based upon work supported by the National Science Foundation
under Grant No. DGE 1106400 (AGA).  Any opinions, findings, and conclusions or
recommendations expressed in this material are those of the author(s) and do
not necessarily reflect the views of the National Science Foundation.  This
work was supported in part by Systems on Nanoscale Information fabriCs (SONIC),
one of the six SRC STARnet Centers, sponsored by MARCO and DARPA (AGA).

%% file: supp_info.tex
\section{Supplementary Information}

\subsection{Expected Angles}
\label{sec:angle_theory}

We draw random $n$
dimensional vectors from a rotationally invariant distribution and compare the
angles between two random vectors and the binarized version of that vector.
We note that a rotationally invariant distribution can be factorized into a pdf
for the magnitude of the vector times a distribution on angles. In the
expectations that we are calculating, the magnitude cancels out and there is
only one rotationally invariant distribution on angles. Thus it suffices to
compute these expectations using a Gaussian.

Lemmas:
\begin{enumerate}
    \item Consider a vector, $v$, chosen from a standard normal distribution of dimension $n$.
        Let $\rho = \frac{v_1}{\sqrt{v_1^2+...+v_n^2}}$. Then $\rho$ is distributed according
to:
$g(\rho) = \frac{1}{\sqrt{\pi}} \frac{\Gamma(n/2)}{\Gamma((n-1)/2)}(1-\rho^2)^{\frac{n-3}{2}}$
\url{http://www-stat.wharton.upenn.edu/~tcai/paper/Coherence-Phase-Transition.pdf}
where $\Gamma$ is the Gamma function.


\item $\frac{\Gamma(z+\alpha)}{\Gamma(z+\beta)} = z^{\alpha-\beta}\left(1 +
    \frac{(\alpha-\beta)(\alpha+\beta+1)}{2z}\right) + O(|z|^{-2})$
    as $z\rightarrow\infty$
\end{enumerate}

\begin{itemize}
    \item Distribution of angles between two random vectors.

Since a Gaussian is a rotationally invariant distribution, we can say without
loss of generality that one of the vectors is $(1, 0, 0,\ldots 0)$. Then the
cosine angle between those two vectors is $\rho$ as defined above. While we
have the exact distribution, we note that

\begin{itemize}
    \item $E(\rho)=0$ due to the symmetry of the distribution.

    \item $Var(\rho) = E(\rho^2) = \frac{1}{n}$ because
        $1 = E\left(\frac{\sum_i x_i^2}{\sum_j x_j^2}\right) = \sum_i
        E\left(\frac{x_i^2}{\sum_j x_j^2}\right) = n * E(\rho^2)$
\end{itemize}

\item Angles between a vector and the binarized version of that vector, $\eta = \frac{v \cdot
\theta(v)}{||v|| \cdot ||\theta(v)||}
=\frac{\sum_i |v_i|}{\sqrt{\sum v_i^2} * \sqrt{n}}$

\begin{itemize}
\item
$$E(\eta) = \frac{\sqrt{n}}{\sqrt{\pi}} * \frac{\Gamma(n/2)}{\Gamma((n+1)/2)}
\qquad \lim_{n\rightarrow\infty} E(\eta) =  \sqrt{\frac{2}{\pi}}$$

First, we note $E(\eta) = \sqrt{n} E(|\rho|)$. Then
$E(|\rho|) = \int_0^1 d\rho \, \rho\frac{2}{\sqrt{\pi}}
\frac{\Gamma(n/2)}{\Gamma((n-1)/2)}(1-\rho^2)^{\frac{n-3}{2}} =
\frac{2}{\sqrt{\pi}} * \frac{1}{n-1}\frac{\Gamma(n/2)}{\Gamma((n-1)/2)}$
(substitute $u=\rho^2$ and use $\Gamma(x+1) = x \Gamma(x)$ ). Lemma two gives
the $n\rightarrow\infty$ limit.

\item
$$Var(\eta) = \frac{1}{n} \left(1 - \frac{1}{\pi}\right) + O(1/n^2)$$
Thus we have the normal scaling as in the central limit theorem of the large
$n$ variance. We can calculate this explicitly following the approach of
\url{https://en.wikipedia.org/wiki/Volume_of_an_n-ball#Gaussian_integrals}.

As we've calculated $E(\eta)$, it suffices to calculate $E(\eta^2)$. Expanding
out $\eta^2$, we get $E(\eta^2) = \frac{1}{n} + (n-1) * E(\frac{|v_1
v_2|}{v_1^2+\ldots v_n^2})$. Below we show that
$E(\frac{|v_1 v_2|}{v_1^2+\ldots v_n^2}) = \frac{2}{\pi n}$. Thus the variance is:

$$\frac{1}{n} * \left(1 - \frac{2}{\pi}\right) + \frac{2}{\pi} - \left(
\frac{\sqrt{n}}{\sqrt{\pi}} \frac{\Gamma(n/2)}{\Gamma((n+1)/2)}
\right)^2
$$

Using Lemma 2 to expand out the last term, we get $[\frac{\sqrt{n}}{\sqrt{\pi}}
(n/2)^{-1/2}(1- 1/(4n) + O(n^{-2}))]^2$ = $\frac{2}{\pi}(1-1/(2n) +
O(n^{-2}))$. Plugging this in gives the desired result.

Going back to the calculation of that expectation, change variables to
$v_1=r\cos\theta$, $v_2 = r\sin\theta$, $z^2 = v_3^2+...+v_n^2$.  The
integration over the volume
element $dv_3\ldots dv_n$ is rewritten as $dz dA_{n-3}$
where $dA_n$ denotes the surface element of a $n$ sphere. Since the
integrand only depends on the magnitude, $z$,
$\int dA_{n-3} = z^{n-3} * S_{n-3}$ where $S_n = \frac{2 \pi^{(n+1)/2}}{\Gamma(\frac{n+1}{2})}$ denotes the surface area of a
unit $n$-sphere. Then

$$E\left(\frac{|v_1 v_2|}{v_1^2+\ldots v_n^2}\right) =
(2\pi)^{-n/2} S_{n-3} \int_0^{2\pi} d\theta |\cos\theta\sin\theta| * \int r dr  z^{n-3} dz *
\frac{r^2}{r^2+z^2} * e^{-(z^2+r^2)/2}$$

Then substitute $r = p \cos\phi$, $z=p\sin \phi$ where $\phi \in [0, \pi/2]$


$$= (2\pi)^{-n/2} * 2 S_{n-3} \int_0^{\pi/2} d\phi \cos\phi ^ 3 * \sin \phi ^{n-3}
\int_0^\infty  dp  * p^{n-1} e^{-p^2/2}$$

The first integral is $\frac{2}{n(n-2)}$ using $u=\sin^2\phi$. The second
integral is $2^{(n-2)/2} \Gamma(n/2)$ using $u=p^2/2$ and the definition of the
gamma function.
Simplifying, we get $\frac{2}{\pi * n}$.
\end{itemize}
\end{itemize}
Roughly speaking, we can see that the angle between a vector and a binarized
version of that vector converges to $\arccos \sqrt{\frac{2}{\pi}} \approx
37^\circ$ which is a very small angle in high dimensions.

\subsection{An Explicit Example of Learning Dynamics}
\label{sec:simple_dynamics}
In this subsection, we look at the learning dynamics for the BNN
training algorithm in a simple case and gain some insight about the learning algorithm.
Consider the case of regression where we try and predict $y$ with $x$ with a
binary linear predictor. Using a squared error loss, we have
$L = (y - \hat{y})^2 = (y - w^b x)^2 = (y - \theta(w^c)x)^2$.
(In this notation, $x$ is a column vector.)
Taking the derivative of this loss with respect to
the continuous weights and using the rule for back propagating through the
binarize function, we get $\Delta w^c \sim - dL/dw^c = - dL/dw^b \cdot
dw^b/dw^c = (y - w^b x) x^T$.
Finally, averaging over the training data, we get

\begin{equation}
    \Delta w^c \sim C_{yx} - \theta(w^c) \cdot C_{xx}
    \qquad C_{yx} = E[yx^T]
    \qquad C_{xx} = E(x x^T)
\end{equation}
It is worthwhile to compare this equation the corresponding equation from
typical linear regression:
$\Delta w^c \sim C_{yx} - w^c \cdot C_{xx}$
For simplicity, lets consider the case where $C_{xx}$ is the identity matrix.
In this case, all of the components of $w$ become independent and we get the
equation
$\delta w = \epsilon * (\alpha - \theta(w))$ where $\epsilon$ is the learning rate and
$\alpha$ is the entry of $C_{yx}$ corresponding to a particular element, $w$.
If we were doing regular linear regression, it is clear that the stable point
of these equations is when $w = \alpha$. Since we binarize the weight, that
equation cannot be satisfied. However, it can be shown $(\star)$ that in this special case of binary weight
linear regression, $E(\theta(w^c)) = \alpha$.

Intuitively, if we consider a high dimensional vector and
the fluctuations of each component are likely to be out
of phase, then $w^b\cdot x \approx w^c \cdot x$ is going to be correct in expectation with a
variance that scales as $\frac{1}{n}$.  During the actual learning process, we anneal the
learning rate to a very small number, so the particular state of a fluctuating
component of the vector is frozen in. Relatedly, the equation $C_{yx} \approx
w C_{xx}$ is easier to satisfy in high dimensions, whereas in low dimensions,
we only satisfy it in expectation.

Rough proof for $(\star)$: Suppose that $|\alpha| \le 1$.   The basic idea of these dynamics is that you are
taking steps of size proportional to $\epsilon$ whose direction depends on
whether $w>0$ or $w<0$. In particular, if $w>0$, then we take a step $-
\epsilon \cdot |1 - \alpha|$ and if $w < 0$, we take a step $\epsilon \cdot
(\alpha + 1)$.  It is evident that after a sufficient burn-in period, $|w| \le
\epsilon * \max(|1-\alpha|, 1+\alpha) \le  2\epsilon$.
Suppose $w>0$ occurs with fraction $p$ and $w<0$ occurs with
fraction $1-p$.  In order for $w$ to be in equilibrium, oscillating about zero,
we must have that these steps balance out on average:
$p (1-\alpha) = (1-p)(1+\alpha)\rightarrow p = (1+\alpha)/2$.
Then the expected value of $\theta(w)$ is $1 * p + (-1) * (1-p) = \alpha$.  When
$|\alpha| > 1$, the dynamics diverge because $\alpha - \theta{bin} (w)$ will
always have the same sign. This divergence demonstrates the importance of some
normalization technique such as batch normalization or attempting to represent
$w$ with a constant times a binary matrix.

\subsection{Dot Product Correlations After Activation Permutation}
\label{sec:perm_corr}

Suppose that we look at $A = w\cdot a$ and $B = v\cdot a$ where $a$ are now the
randomly permuted activations. What does the distribution of $A, B$ look like?
To answer this, we look at the correlation between $A$ and $B$ and show that it
is the correlation between $w$ and $v$. First, let us assume that $p(a) =
\prod_i f(a_i)$ $E(a_i) = 0$, $E(a_i^2) = \sigma^2$. Then $E(A)=E(B)=0$. Now we compute:

$E(AB) = \sum_{i, j} w_i v_j E(a_ia_j) = \sigma^2 (w\cdot v)$
Likewise, $E(A^2) = \sigma^2 (w\cdot w)$ and $E(B^2) = \sigma^2 (v\cdot v)$. Thus
the correlation coefficient $\frac{w\cdot v}{|w||v|}$, as desired

\listoffixmes




